\documentclass[acmlarge]{acmart}

\usepackage{graphicx}
\usepackage{color}
\usepackage{booktabs}
\usepackage{caption}
\usepackage{subcaption}
\usepackage{multirow}
\usepackage{url}

\AtBeginDocument{%
  }

\begin{document}

\title{Adversarial Pretraining of Self-Supervised Deep Networks: Past, Present and Future}

\author{Guo-Jun Qi}
\email{guojunq@gmail.com}
\affiliation{%
  \institution{Laboratory for Machine Perception and Learning}
  \streetaddress{10940 NE 33RD PLACE, SUITE 202}
  \city{Bellevue}
  \state{Washington}
  \country{USA}
  \postcode{98004}
}

\author{Mubarak Shah}
\affiliation{%
  \institution{University of Central Florida}
  \streetaddress{4328 Scorpius St.}
  \city{Orlando}
  \state{Florida}
  \country{USA}
  \postcode{32816}}
\email{shah@crcv.ucf.edu}
%
%
%
%
%
%

\renewcommand{\shortauthors}{Qi and Shah}

\begin{abstract}
In this paper, we review adversarial pretraining of self-supervised deep networks including both convolutional neural networks and vision transformers.  Unlike the adversarial training with access to labeled examples, adversarial pretraining is complicated as it only has access to unlabeled examples. To incorporate adversaries into pretraining models on either input or feature level, we find that existing approaches are largely categorized into two groups: memory-free instance-wise attacks imposing worst-case perturbations on individual examples, and memory-based adversaries shared across examples over iterations. In particular, we review several representative adversarial pretraining models based on Contrastive Learning (CL) and Masked Image Modeling (MIM), respectively, two popular self-supervised pretraining methods in literature. We also review miscellaneous issues about computing overheads, input-/feature-level adversaries, as well as other adversarial pretraining approaches beyond the above two groups. Finally, we discuss emerging trends and future directions about the relations between adversarial and cooperative pretraining, unifying adversarial CL and MIM pretraining, and the trade-off between accuracy and robustness in adversarial pretraining.
\end{abstract}


\maketitle

\section{Introduction}\label{sec:intro}
Adversarial pretraining aspires to learn an unsupervised deep networks without access to labels. In contrast, adversarial training in literature \cite{tsipras2018robustness,kurakin2016adversarial,goodfellow2014explaining,szegedy2013intriguing,madry2017towards,moosavi2016deepfool,papernot2016distillation,carlini2017towards,xu2017feature,zhang2020wcp} seeks to find worst-case adversarial examples and use them to train neural networks robust to the corresponding attacks.  While some findings \cite{szegedy2013intriguing,tsipras2018robustness} revealed that the adversarially trained networks can be robust to adversarial attacks or gain improved standard accuracy, however, there are limited reviews of {\em adversarially pretrained networks} by classifying and evaluating existing approaches, assessing their advantages and shortcomings, as well as charting future directions.

At the start, we want to clarify a common misunderstanding about the role of adversarial pretraining.  The goal of an adversarial approach is not limited to learning robust representation against potential attacks. Instead, it is also employed to improve the generalization accuracy in downstream tasks, especially when the adversarial model attacks on feature levels \cite{hu2021adco,robinson2021can,robinson2020contrastive,kalantidis2020hard} rather than on raw inputs (e.g., image pixels) of individual instances (aka {\em instance-wise attacks} \cite{ho2020contrastive,kim2020adversarial,jiang2020robust}). When not attacking on the raw inputs, the adversarial pretraining often cares about whether the learned {\em features} are generalizable to future problems, aiming to avoid learning trivial solutions that merely use low-level features to bypass a pretext task. For example, easy negatives in contrastive learning could result in less discriminative features to distinguish between positive and negative samples for a query \cite{kalantidis2020hard,hu2021adco}; in masked imaging modeling (MIM), the network may learn low-level features to reconstruct missing patches by simply using the similarity between locally correlated patches \cite{shi2022adversarial,anonymous2022adpe} if the MIM objective is not sufficiently hard. In these cases, it is beneficial to explore adversarial approaches to improve the generalizability of learned representations. In other words, learning more generalizable representations through adversarial pretraining is an equally important goal in literature as learning robust representations against some presumptive attacks.

\subsection{Instance-wise Perturbations from Adversarial Training to Pretraining}
Let us begin by revisiting the adversarial training. Formally, the adversarial training (instead of pretraining) gets access to the labeled examples, and it maximizes an associated classification loss such as the cross entropy $\mathcal L_{ce}$ over a constraint perturbation $\boldsymbol \delta$ of  magnitude $\varepsilon$ to find an adversarial example for an input $\mathbf x$ \cite{madry2017towards}, i.e.,
\begin{equation}\label{eq:ad_ex}
\boldsymbol \delta^{\star} = \arg\max\limits_{\|\boldsymbol\delta\|_p\leq\varepsilon}\mathcal L_{ce} (\mathbf x + \boldsymbol \delta, \mathbf y; \boldsymbol\theta),
\end{equation}
where the network weights $\boldsymbol\theta$ can be learned by
\begin{equation}\label{eq:ad_tr}
\boldsymbol\theta^{\star} = \arg\min\limits_{\boldsymbol\theta} \mathop\mathbb E_{(\mathbf x, \mathbf y)\sim\mathcal D}\mathcal L_{ce}~(\mathbf x + \boldsymbol \delta^{\star}, \mathbf y; \boldsymbol\theta)
\end{equation}
with the expectation taken over labeled examples $(\mathbf x, \mathbf y)$ sampled from a distribution $\mathcal D$. More detailed reviews of adversarial training of deep networks can be found in \cite{akhtar2018threat,akhtar2021advances}.

Instead, the adversarial pretraining aligns with unsupervised representation learning without access to label $\mathbf y$ as in the above formulation \cite{wu2018unsupervised,chen2020simple,agrawal2015learning,gidaris2018unsupervised,qi2019learning,doersch2015unsupervised,kim2018learning,larsson2016learning,zhang2019aet,qi2020small}, and has separate pretext and downstream tasks that are often more complex than the adversarial training. First, there are many different ways to define a pretext task to unsupervisedly learn a deep network.  A variety of pretext tasks result in different approaches to perform meaningful adversarial pretraining, and some pretext tasks may make the pretrained representation vulnerable to potential attacks \cite{gupta2022higher}. In this review, we will show that a straight extension of adversarial training is to add perturbations on unlabeled examples and solve a similar minimax problem for adversarial pretraining.  Particularly, among popular pretext tasks are Contrastive Learning (CL) \cite{oord2018representation,henaff2019data,wu2018unsupervised,chen2020improved,chen2020simple} and Masked Image Modeling (MIM) \cite{he2022masked,xie2022simmim,bao2021beit} for Convolutional Nets (ConvNet) \cite{krizhevsky2012imagenet,he2016deep} and vision transformers \cite{dosovitskiy2020image,touvron2021training,wang2022dual}, respectively.

We will review how adversarial perturbations can be generalized to these pretext tasks on either instance or feature level. Particularly, we will review the related works from a novel perspective by grouping them into memory-free and memory-based adversarial pretraining. We will show that the memory-free adversarial approaches usually consist of instance-wise perturbations as in the adversarial training methods, where the perturbation is constructed on raw inputs of individual examples. Pretraining a deep network is memory-free in the sense that the perturbations and the associated adversarial examples are not kept over epochs.  Many adversarial approaches belong to this category including \cite{ho2020contrastive,kim2020adversarial,jiang2020robust}.

\subsection{Feature-level adversarial Pretraining with Memory-based Adversaries}
In contrast, the feature-level adversarial pretraining abandons construction of instance-wise perturbations, and instead uses a shared memory as adversarial players. A natural choice of such adversaries is the memory bank widely used in contrastive learning \cite{he2020momentum},  and thus most of feature-level adversarial pretraining is memory-based. For example, AdCo \cite{hu2021adco} treats all examples in the memory bank (cf. \cite{he2020momentum}) as {\em learnable} negatives, and directly learns them by maximizing the contrastive loss. This results in hard negatives that are continuously updated to be mixed with their positive counterparts. Hence, the learned negatives are hard to be distinguished from positives so more discriminative representations must be learned that are generalizable to downstream tasks. IFM \cite{robinson2021can} extends the idea of AdCo by considering implicit feature-level modification to contrastive pairs subject to assigned budgets.
These methods give rise to a family of feature-level adversarial learning to generate hard negatives in a principled manner, establishing strong connections with hard negative mining and sampling methods \cite{kalantidis2020hard,robinson2020contrastive} that we will also review in this survey.

\subsection{MIM-based Adversarial Pretraining}
Moreover, Masked Image Modeling (MIM) \cite{he2022masked,xie2022simmim,bao2021beit} has attracted increasing attentions for pretraining vision transformers.  While most of existing adversarial pretraining methods are built upon contrastive learning, this opens up unprecedented research opportunities to study adversarial pretraining of vision transformers in the MIM framework. We will review existing adversarial MIM-pretraining approaches for transformers \cite{shi2022adversarial,anonymous2022adpe}, and point out some natural extension of adversarial perturbations to the MIM. Both instance-wise (e.g., FGSM-based approach) and feature-level adversarial MIM-pretraining approaches will be reviewed and discussed.

\subsection{Future Directions and Survey Structure}
Finally, we will review some emerging trends and future directions on adversarial pretraining. In particular, we will focus on three aspects of future directions.
\begin{itemize}
\item {\bf Adversarial vs. cooperative pretraining.} We will review related works and discuss the connection between adversarial (that maximizes the training loss) and cooperative (that minimizes the training loss) pretraining. We will point out a direction to study when different parts of a pretrained network ought to be learn adversarially or cooperatively. In particular,  depending on different modes of adversarial pretraining (instance-wise vs. memory-based, and input-level vs. feature-level), a positive query in contrastive learning can be treated as either an adversary in a hybrid model or a cooperator through a shared memory bank. We will show that it could be beneficial to combine these two modes through various pretraining approaches.

\item {\bf Unifying adversarial contrastive learning and masked image modeling.} Existing pretraining approaches \cite{zhou2021ibot,huang2022contrastive} have shown that combining contrastive learning and masked image modeling leads to more powerful representation of vision transformers with improved generalizability. The second direction worthy of studying is to explore an elegant way of designing adversaries to integrate both approaches. It is expected that the unified adversarial pretraining can make the learned representation more powerful and generalizable to downstream tasks.

\item {\bf Accuracy vs. robustness.} In adversarial training, the relationship between standard accuracy and model robustness to adversarial perturbations has been extensively studied.  Existing works have demonstrated that the network robustness may not always be indicative of improved accuracy \cite{tsipras2018robustness}. The problem becomes more complicated for adversarial pretraining, since the pretext tasks used to pretrain a network are usually not related to the downstream objectives. Revealing the underlying connections between network robustness to pretext task adversaries and generalization accuracy in downstream tasks should be a direction worthy of exploring from both theoretical and practical perspectives.
\end{itemize}

The remainder of this paper is organized as follows. In Section~\ref{sec:acl}, we will review the contrastive pretraining methods by grouping them into instance-wise memory-free and feature-level memory-based models. We will discuss some miscellaneous issues regarding the computing overheads, and combined objectives for pretraining.  In Section~\ref{sec:amim}, we will review the MIM-based adversarial pre-training, followed by other related methods beyond contrastive and MIM pretraining in Section~\ref{sec:other}. We will discuss the evaluation protocols and review existing results in Section~\ref{sec:eval}.
Emerging trends and future directions will be discussed in Section~\ref{sec:emerging}, and we will conclude the paper in Section~\ref{sec:concl}.

\begin{table*}[h]
\caption{Different types of adversarial pre-training models discussed in the paper. For each model, we denote its model type (memory-free vs. memory-based), base model(CL, MIM, or others), the formula for adversaries, pretraining adversaries on input or feature levels (Input/Feature), and if using a pre-training objective combining adversarial and standard losses (in the column ``Comb. obj." ). The table compares the similarities and differences across different models in forms of the formula for adversaries.}\label{tab:models}\vspace{-2mm}
\centering
\small
\setlength{\tabcolsep}{1mm}{
 \begin{tabular}{l|ccccc} \toprule
Model& Type &Base&Adversaries&Input/Feature&Comb. obj.\\ \midrule\midrule
RoCL~\cite{kim2020adversarial}&\multirow{3}{*}{memory-free}&\multirow{3}{*}{CL}& \multirow{3}{*}{$\begin{aligned} &t(\mathbf x)\leftarrow t(\mathbf x)~~~~~~~~~~~~~~~~~~~~~\qquad\qquad\cdots{\rm FGSM}\\&+\epsilon~{\rm sign}(\nabla_{t(\mathbf x)}\mathcal L_{con} (t(\mathbf x), \{\mathbf x_{pos}\},\{\mathbf x_{neg}\}) )\\&\qquad\qquad\qquad\qquad{\rm or}\\&t(\mathbf x) \leftarrow \mathop\Pi\limits_{\mathcal S(t(\mathbf x),\epsilon)} [t(\mathbf x) \quad\quad~~~~~~~~\cdots{\rm PGD}\\
&+ \alpha{\rm sign}(\nabla_{t(\mathbf x)}\mathcal L_{con} (t(\mathbf x), \{\mathbf x_{pos}\},\{\mathbf x_{neg}\}))]\end{aligned}$} &\multirow{3}{*}{input-level}&\multirow{3}{*}{yes}\\\\\\CLAE~\cite{ho2020contrastive}\\\\\\ACL~\cite{jiang2020robust}\\\midrule
ADVCL~\cite{fan2021does}&memory-free&CL&$\begin{aligned}&\mathbf x\leftarrow \mathbf x+\epsilon~{\rm sign}(\nabla_{\mathbf {\bar x}}\mathcal L_{con}(t(\mathbf x), t'(\mathbf x), \mathbf{\bar x},\mathbf x_h))\\&{\rm by~~FGSM~~ attacks.}\end{aligned}$&input-level&yes\\\\
ARoCL~\cite{gupta2022higher}&\multirow{2}{*}{memory-free}&\multirow{2}{*}{CL}&\multirow{2}{*}{false negative removal} & \multirow{2}{*}{input-level}&\multirow{2}{*}{yes}\\\\AACL~\cite{gupta2022higher}\\\midrule
AdCo~\cite{hu2021adco}&memory-based&CL&$\mathbf z_{neg} \leftarrow \mathbf z_{neg} + \dfrac{\alpha}{\tau}\mathop \mathbb E\limits_{\mathbf x\sim \mathcal D} p(\mathbf z_{neg}|\mathbf z)~\mathbf z$&feature-level&no\\\\
CaCo~\cite{wang2022caco}&memory-based&CL&$\mathbf z_{pos} \leftarrow \mathbf z_{pos} + \dfrac{\alpha}{\tau}[1-p(\mathbf z_{pos}|\mathbf z)] \mathbf z$&feature-level&no\\\midrule
AdPE~\cite{anonymous2022adpe}&{memory-based}&{MIM}&\multirow{3}{*}{$\begin{aligned}&\boldsymbol\delta\leftarrow\mathop\Pi\limits_{\|\boldsymbol\delta\|_q\leq \epsilon}[\boldsymbol\delta+\alpha\nabla_{\boldsymbol\delta}\cdot~~~~~~~~~~~~~~~\qquad\cdots{\rm PGD}\\
&\mathcal L_{MIM}(\{\mathbf p_{[\widetilde g(x_i-x_j+\delta_x),\widetilde g(y_i-y_j+\delta_y)]}|i\in\mathbb M\};\boldsymbol\theta)]\end{aligned}$}&feature-level&no\\\\\\\midrule
IFM~\cite{robinson2021can} & memory-free & CL &$\begin{aligned}
\mathbf z_{pos} \leftarrow \mathbf z_{pos} - \epsilon \mathbf z\\
\mathbf z_{neg} \leftarrow \mathbf z_{neg} + \epsilon \mathbf z
\end{aligned}$ &feature-level&yes\\\midrule
MoCHi~\cite{kalantidis2020hard}&--&CL&hard negative mixing&feature-level&yes\\\\
HCL~\cite{robinson2020contrastive}&--&CL&hard negative sampling&feature-level&yes\\\midrule
BYORL~\cite{gowal2020self}&memory-free&BYOL&$\boldsymbol\delta \leftarrow \Pi_{\|\boldsymbol\delta\|_p\leq \epsilon}[\boldsymbol \delta + \alpha \nabla_{\boldsymbol\delta} \mathcal L_{byol}(t(\mathbf x)+\boldsymbol\delta, t'(\mathbf x))]$&input-level&no\\\\\midrule
RUSH~\cite{pang2022rush}&memory-free&CL&randomized smoothing&input-level&no\\\midrule
ADIOS~\cite{shi2022adversarial}&memory-free&MIM&mask generating network $\mathcal M_\phi$&input-level&no\\\bottomrule
\end{tabular}}
\begin{flushleft}
Remarks: \\
1) Memory-free: instance-wise perturbations imposed on individual examples without carrying over iterations;\\
2) Memory-based: adversarial samples are stored in a shared memory to challenge the pretrained network;\\
3) Input-level: adversaries imposed on raw inputs of instances;\\
4) Feature-level: adversaries imposed on feature representations.
\end{flushleft}
\end{table*}

\begin{table*}[th!]
\caption{A summary of source code links.}\label{tab:source}
\centering
 \begin{tabular}{l|l} \toprule
Models& Links\\ \midrule\midrule
RoCL~\cite{kim2020adversarial} & \url{https://github.com/Kim-Minseon/RoCL}\\
CLAE~\cite{ho2020contrastive} & \url{https://github.com/chihhuiho/CLAE}\\
ACL~\cite{jiang2020robust} & \url{https://github.com/VITA-Group/Adversarial-Contrastive-Learning}\\
AdCo~\cite{hu2021adco} & \url{https://github.com/maple-research-lab/AdCo}\\
CaCo~\cite{wang2022caco} & \url{https://github.com/maple-research-lab/CaCo}\\
IFM~\cite{robinson2021can} & \url{https://github.com/joshr17/IFM}\\
MoCHi~\cite{kalantidis2020hard} & \url{https://europe.naverlabs.com/research/computer-vision/mochi/}\\
HCL~\cite{robinson2020contrastive}&\url{https://github.com/joshr17/HCL}\\
ADIOS~\cite{shi2022adversarial}&\url{https://github.com/yugeten/adios}\\
ADVCL~\cite{fan2021does}&\url{https://github.com/LijieFan/AdvCL}
\\\bottomrule
\end{tabular}
\end{table*}

\section{Adversarial Contrastive Pretraining}\label{sec:acl}

\begin{figure*}[t]
    \centering
    \begin{subfigure}[c]{1.0\textwidth}
        \includegraphics[width=\textwidth]{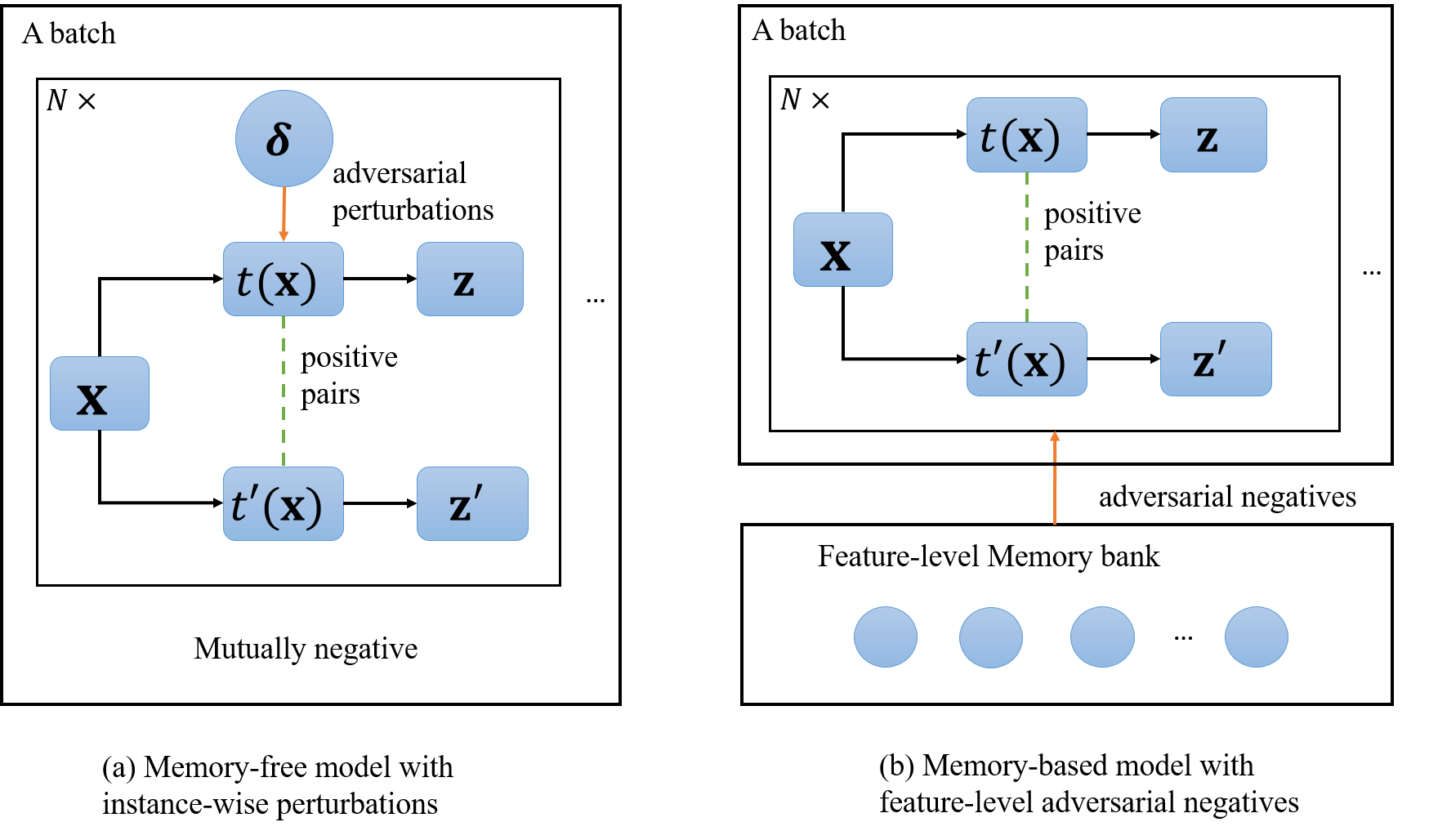}
    \end{subfigure}\\
    ~ 
    \caption{Instance-wise memory-free vs. feature-level memory-based adversarial pretraining models. At each iteration, a batch of $N$ instances are sampled (represented in the figure by the rectangle box denoted by ``A batch"). Given an input instance $\mathbf x$, its two augmented views $t(\mathbf x)$ and $t'(\mathbf x)$ with two transformations $t$ and $t'$ randomly drawn from $\mathcal T$ are created. The two views are encoded and projected into two embedded feature representations $\mathbf z$ and $\mathbf z'$, respectively. The two augmented views and their representations form a positive pair (dashed green line).  (a) For memory-free model with instance-wise perturbations, an optimal perturbation $\boldsymbol \delta$ is computed for each individual sample $t(\mathbf x)$ in the batch by maximizing the contrastive loss. Different samples in the batch are considered as mutually negative to one another. This adversarial pretraining model is memory-free as the perturbation attacks are sample-specific and they will not be carried to the next iteration. (b) For a memory-based model with adversarial negatives, there exists a memory bank of adversarial negatives that are shared among all query samples in a batch.  In other words, for every query sample $t(\mathbf x)$ or $t'(\mathbf x)$ in the batch, all adversarial negatives in the memory bank are considered as being negative to it. These negatives are adversarially learned by maximizing the contrastive loss. This model is memory-based since the adversarial negatives in the memory bank are kept over iterations, and are continuously updated over time. In both figures, we use the orange arrow to highlight the adversarial component in the two models.}\label{fig:memory}
\end{figure*}

Contrastive learning has become the state-of-the-art method for pretraining a variety of deep networks ranging from convolutional networks \cite{chen2020improved,he2020momentum} to vision transformers \cite{caron2021emerging}. Incorporating adversaries into contrastive learning for pretraining deep networks has also been intensively studied in literature \cite{ho2020contrastive,kim2020adversarial,jiang2020robust,hu2021adco}.

In this section, we will review existing methods from a novel perspective by categorizing them into two large groups: memory-free methods with instance-wise perturbations, memory-based methods on feature level by learning a shared memory of adversarial negatives.  
Readers may take a glance at Table~\ref{tab:models} that summarizes different adversarial pretraining models we will discuss in this paper.


\subsection{Background: Contrastive Learning}

Contrastive learning seeks to pre-train an encoder network by maximizing the agreement of representations between a pair of samples transformed from the same instance, while pushing apart the representations of those transformed from different ones \cite{wu2018unsupervised}.

Formally, given an instance $\mathbf x$, it is randomly transformed into a pair of examples $t(\mathbf x)$ and $t'(\mathbf x)$ with two transformations $t$ and $t'$ drawn from a distribution $\mathcal T$. An encoder network  $f_\theta$ and a following projector $g_\theta$ map the pair into two latent vectors $\mathbf z = g_\theta(f_\theta(t(\mathbf x)))$ and $\mathbf z' = g_\theta(f_\theta(t'(\mathbf x)))$, respectively.  Then, the contrastive learning is designed to minimize the following contrastive loss
\begin{equation}
\begin{aligned}
\mathop \mathbb E\limits_{\mathbf x\sim \mathcal D} &\mathcal L_{con} (t(\mathbf x), \{\mathbf x_{pos}\},\{\mathbf x_{neg}\};\theta)
&\triangleq -\log \dfrac{\mathop\sum\limits_{\mathbf z'\in\{\mathbf z_{pos}\}}\exp(sim(\mathbf z,\mathbf z')/\tau)}{\mathop\sum\limits_{\mathbf z''\in\{\mathbf z_{pos}\}\cup\{\mathbf z_{neg}\}}\exp(sim(\mathbf z,\mathbf z'')/\tau)},
\end{aligned}
\end{equation}
where $\mathcal D$ is the data distribution, $\{\mathbf x_{pos}\}$ and $\{\mathbf x_{neg}\}$ are the set of positive and negative examples for $t(\mathbf x)$, and $\{\mathbf z_{pos}\}$ and $\{\mathbf z_{neg}\}$ are their representations. Usually, the positive set has a single example  $t'(\mathbf x)$  transformed from the same instance, and the negative set contains all examples transformed from different instances. The $sim$ is a similarity function usually chosen as cosine similarity, and $\tau$ is the temperature for the loss.

\subsection{Memory-free Methods: Instance-wise Perturbation}\label{sec:mfree}

The idea of instance-wise attack against the contrastive pre-training is to seek a worst-case perturbation $\boldsymbol \delta$ onto the transformed example $t(\mathbf x)$ by maximizing the above contrastive loss, yielding the most dissimilar positive sample $t(\mathbf x)+\boldsymbol \delta$. Then, the contrastive loss is minimized by maximizing the agreement between the adversarially perturbed example and its clean counterparts in $\{\mathbf x_{pos}\}$, leading to a minimax problem to pre-train a deep network parameterized with $\theta$,
\begin{equation}\label{eq:mfree}
\min_{\theta} \mathop \mathbb E\limits_{\mathbf x\sim\mathcal D} \max_{\|\boldsymbol\delta\|_p \leq \epsilon} \mathcal L_{con} (t(\mathbf x)+\boldsymbol\delta, \{\mathbf x_{pos}\},\{\mathbf x_{neg}\};\theta),
\end{equation}
where the perturbation magnitude is constrained by an $\ell_p$-norm of $\epsilon$.

The inner maximization problem can be solved by the Fast Gradient Sign Method (FGSM) \cite{goodfellow2014explaining} for an infinity norm $\ell_{+\infty}$,
\begin{equation}\label{eq:mfree_delta}
\boldsymbol\delta^\star = \epsilon~{\rm sign}(\nabla_{t(\mathbf x)}\mathcal L_{con} (t(\mathbf x), \{\mathbf x_{pos}\},\{\mathbf x_{neg}\};\theta)).
\end{equation}
Or the Projected Gradient Descent (PGD) \cite{madry2017towards} can be applied to iteratively update $\boldsymbol\delta$ to find an optimal perturbation,
\begin{equation}
\begin{aligned}
t(\mathbf x) \leftarrow \mathop\Pi\limits_{\mathcal S(t(\mathbf x),\epsilon)} [t(\mathbf x)
+ \alpha{\rm sign}(\nabla_{t(\mathbf x)}\mathcal L_{con} (t(\mathbf x), \{\mathbf x_{pos}\},\{\mathbf x_{neg}\};\theta))],
\end{aligned}
\end{equation}
with a step size $\alpha$ and a projection $\Pi$ onto the constraint $\mathcal S(t(\mathbf x),\epsilon)=\{t(\mathbf x)+\boldsymbol\delta|\|\boldsymbol\delta\|_p\leq\epsilon\}$.

The sought perturbation $\boldsymbol\delta$ is distinctive to an individual sample $t(\mathbf x)$, given the current estimate of network weights $\theta$.  For each iteration, a new batch of samples arrive, and the adversarial positives $t(\mathbf x)+\boldsymbol \delta$ from the past iterations are abandoned since the network weights are updated based on the adversarial examples of the new batch. In this sense, such instance-wise attacks are {\em memory-free} with no need to store the past adversarial examples.

In addition to such vanilla instance-wise perturbations \cite{kim2020adversarial}, there are several variants. For example, \cite{jiang2020robust} considered to apply a pair of perturbations $\boldsymbol \delta$ and $\boldsymbol \delta'$ to both positive samples $t(\mathbf x)$ and $t'(\mathbf x)$. On the other hand, \cite{ho2020contrastive} only used a transformed sample $t(\mathbf x)$ and its adversarial perturbation $t(\mathbf x)+\boldsymbol\delta$ to form a positive pair, without involving another transformed sample $t'(\mathbf x)$ from the same instance $\mathbf x$.

{\noindent \bf More variants.} Besides the above two, an asymmetric InfoNCE (A-InfoNCE) loss \cite{yu2022adversarial} was proposed to mitigate the instance-level identity confusion arising from strong adversarial examples by adopting an asymmetric similarity and re-weighting hard negatives following \cite{robinson2020contrastive,chuang2020debiased}.  In \cite{gupta2022higher}, it shows that self-supervised models are more susceptible to adversarial perturbations than supervised ones due to false negatives, and hence a false negative removal strategy was proposed to augment RoCL \cite{kim2020adversarial} and ACL \cite{jiang2020robust} in an effort to make the adversarially pretrained representations more robust against attacks.


In \cite{fan2021does}, optimal adversarial perturbations are sought over input images rather than transformed views in defense to adversarial attacks on original inputs. Formally, with an additional high-frequency view $\mathbf x_h$ of an input image $\mathbf x$ through Fourier transformation, the multi-view contrastive pre-training was performed by
$$
\min_\theta \mathop\mathbb E\limits_{\mathbf x\sim\mathcal D}\max_{\|\boldsymbol\delta\|_\infty \leq \epsilon}\mathcal L_{con}(t(\mathbf x), t'(\mathbf x), \mathbf x + \boldsymbol\delta,\mathbf x_h;\theta)
$$
over a contrastive loss $\mathcal L_{con}$ with four views: two transformed view $t(\mathbf x)$ and $t'(\mathbf x)$, the original input $\mathbf x$, and the high-frequency view $\mathbf x_h$. They mutually form the positive pairs for the same $\mathbf x$, while the views from different inputs constitute negative pairs (to avoid notation clutters, we suppress the notations of negative samples in the above loss). Besides such multi-view contrastive learning, the clustering-induced pseudo-labels were treated as the true labels in conventional adversarial training \cite{madry2017towards} to generate adversaries. The resultant pseudo-label loss is combined with the multi-view contrastive loss for adversarial pretraining.

\subsection{Memory-based Methods: Feature-Level Adversarial Negatives}\label{sec:mem}


The Momentum Contrast (MoCo) \cite{he2020momentum} is a memory-based contrastive learning model. While the SimCLR \cite{chen2020simple} treats other instances over a batch as negatives, the MoCo instead uses a {\em memory bank} to store negatives queued from the past iterations.

Built upon such a memory model, the Adversarial Contrast (AdCo) \cite{hu2021adco} treats the negatives in the memory bank as {\em learnable} weights, and uses them as adversaries to attack against the pre-trained network. The memory bank only stores the feature representation $\{\mathbf z_{neg}\}$ of negatives without access to the original samples, and thus the AdCo is also a feature-level adversarial approach. AdCo employs the following minimax objective
$$
\min_{\theta} \max_{\{\mathbf z_{neg}\}} \mathop \mathbb E\limits_{\mathbf x\sim\mathcal D}  \mathcal L_{con} (t(\mathbf x), \{\mathbf x_{pos}\},\{\mathbf z_{neg}\};\theta),
$$
where $\{\mathbf x_{neg}\}$ in (\ref{eq:mfree}) has been replaced with the feature-level representations $\{\mathbf z_{neg}\}$ of negatives for clarity.

Unlike the instance-wise attack relying on the back-propagated gradient to compute the optimal perturbation, one advantage of the memory-based AdCo is that the derivative of the contrastive loss over a negative representation $\mathbf z_{neg}$ can be directly computed \cite{hu2021adco} as
$$
\dfrac{\partial\mathcal L_{con} (t(\mathbf x), \{\mathbf x_{pos}\},\{\mathbf z_{neg}\};\theta)}{\partial \mathbf z_{neg}} =\dfrac{1}{\tau}\mathop \mathbb E\limits_{\mathbf x\sim \mathcal D} p(\mathbf z_{neg}|\mathbf z)~\mathbf z,
$$
where $\mathbf z$ is the projected representation of a transformed sample $t(\mathbf x)$, i.e., $\mathbf z = f_\theta (g_\theta (t(\mathbf x)))$, and
$$
p(\mathbf z_{neg}|\mathbf z) = \dfrac{\exp(sim(\mathbf z,\mathbf z_{neg})/\tau)}{\mathop\sum\limits_{\mathbf z''\in\{\mathbf z_{neg}\}\cup\{\mathbf z'\}}\exp(sim(\mathbf z,\mathbf z'')/\tau)}
$$
where $\mathbf z'$  is the projected representation of another sample $t'(\mathbf x)$ transformed from the same instance. Obviously, $p(\mathbf z_{neg}|\mathbf z)$ is the probability of a negative $\mathbf z_{neg}$ being consistent with the positive anchor $\mathbf z$.

With these derivatives, the representation of a negative sample can be continuously updated given a step size of $\alpha$
\begin{equation}\label{eq:mfree_neg_update}
\mathbf z_{neg} \leftarrow \mathbf z_{neg} + \dfrac{\alpha}{\tau}\mathop \mathbb E\limits_{\mathbf x\sim \mathcal D} p(\mathbf z_{neg}|\mathbf z)~\mathbf z.
\end{equation}
This shows that the memory-bank negatives are pushed towards the representation of positive examples, making it harder to distinguish them from their positive anchors.

The weight $p(\mathbf z_{neg}|\mathbf z)$ in the above update rule implies a self-boosted mechanism -- the more similar the negative $\mathbf z_{neg}$ is to a positive anchor $\mathbf z$, the closer it will be pushed to it. Such a self-boosted mechanism makes it more and more challenging to distinguish the negatives from the positive anchors as they are being mixed together.

The AdCo is a {\em memory-based pretraining} model, since its feature-level memory bank of adversarial negatives are shared over iterations. This is in contrast to the instance-wise perturbations that are distinctive to individual samples in memory-free models. Figure~\ref{fig:memory} compares these two different types of adversarial models.

\subsection{Discussions}

In this section, we elaborate on more subtle aspects to compare the memory-free and memory-based adversarial approaches.

\subsubsection{Computing Overheads: Two Passes vs. One Pass}

In memory-free methods \cite{ho2020contrastive,kim2020adversarial,jiang2020robust}, the instance-wise perturbation $\boldsymbol\delta^\star$ for an individual sample is often computed via FGSM or PGD. This means that at least one forward-and-backward pass is required for FGSM to compute the sample gradient for obtaining the optimal perturbation (i.e., solving the inner maximization in (\ref{eq:mfree})), before another pass is applied to solve the weight gradient to update the network (i.e., solving the outer minimization in (\ref{eq:mfree})).  In other words, at least two passes through a deep network are required to adversarially pretrain a deep network for each iteration.

In comparison, memory-based methods such as AdCo directly update the adversarial negatives without feeding inputs through the network.  Only a single forward-and-backward pass is needed to update the network in an iteration.  This saves the computing cost by half, which does not incur the computing overhead over the original contrastive learning algorithm \cite{hu2021adco}.

\subsubsection{Combining Adversarial and Regular Pre-training}

In the memory-free case, the  adversarial contrastive loss with the instance-wise perturbation is usually combined with the stand contrastive loss with clean examples. In this sense, the adversarial contrastive loss is used as a regularizer to the standard contrastive loss. Experiment \cite{jiang2020robust} showed that the  pre-training combining two losses attains better results.

On the contrary, the memory-based AdCo \cite{hu2021adco} is trained as a stand alone loss over learned adversarial negatives without being combined with the regular contrastive loss.  In this sense, the AdCo model is not trained to be robust against instance-wise attacks. Instead, it focuses on achieving better generalization accuracies in downstream tasks by avoiding trivial solutions with adversarial pretraining on pretext tasks.

\subsubsection{Instance-wise Attacks with Feature-level Adversaries}


Implicit Feature Modification (IFM) \cite{robinson2021can} makes feature-level attacks by adding perturbations to the representations of examples in a batch. This is aligned with the AdCo \cite{hu2021adco} with the feature-level adversarial negatives in the memory bank.

However, unlike the AdCo, the IFM is memory-free instance-wise attacks, whose perturbations are made to individual instances. Specifically, given the representation $\mathbf z$ of an anchor $\mathbf x$, its positive $\mathbf z_{pos}$ and negative $\mathbf z_{neg}$ from a batch are updated by
\begin{equation}
\mathbf z_{pos} \leftarrow \mathbf z_{pos} - \epsilon \mathbf z,
\end{equation}
and
\begin{equation}\label{eq:ifm_neg_update}
\mathbf z_{neg} \leftarrow \mathbf z_{neg} + \epsilon \mathbf z,
\end{equation}
where $\epsilon$ is the magnitude constraint of the perturbation.


The update (\ref{eq:ifm_neg_update}) of IFM pushes a negative representation towards its assigned positive anchor $\mathbf z$.  On the contrary, AdCo applies the update (\ref{eq:mfree_neg_update}) to a memory-bank negative that is pushed towards a batch of positive samples weighted by the probability $p(\mathbf z_{neg}|\mathbf z)$, and AdCo shares the memory-bank negatives for positive anchors in a batch. By sharing negatives among those anchors, the negatives can be updated more efficiently towards the most likely positive counterparts weighted by $p(\mathbf z_{neg}|\mathbf z)$, without being hard assigned to a specific anchor. The experiments \cite{hu2021adco} also verified this by demonstrating that competitive results can be achieved over fewer epochs with a smaller batch.

Finally, compared with the standard instance-wise perturbations on raw inputs, the feature-level adversaries including AdCo and IFM are more computationally efficient.  In particular, with high-resolution images, perturbations on raw pixels could incur expensive computing cost on a large-scale dataset such as Imagenet. This is perhaps why most of existing works \cite{ho2020contrastive,kim2020adversarial,jiang2020robust} on instance-wise perturbations are only evaluated on smaller-scale lower-resolution image datasets, such as CIFAR-10, CIFAR-100 and STL. In this sense, the feature-level adversaries are more amenable to large-scale datasets with high-resolution images.

\section{Adversarial Masked Image Modeling}\label{sec:amim}
In this section, we discuss the emerging problem of pretraining vision transformers via Masked Image Modeling, and review how the adversarial approaches have been proposed for adversarially pretraining these transformers.

Note that the contrastive learning has been used to pretrain vision transformers \cite{chen2021empirical,dosovitskiy2020image}. Thus, there is no difficulty in extending the aforementioned contrastive pretraining to vision transformers. However, due to its natural connection with the pretraining of language transformers \cite{devlin2018bert,brown2020language}, the Masked Image Modeling (MIM) \cite{he2022masked,xie2022simmim,bao2021beit} has attracted extensive attentions recently to pre-train the vision transformers.

In this section, we begin by briefly revisiting the MIM, followed by review of adversarial MIM-pretraining approaches.

The MIM seeks to pre-train a deep network with an inpainting pretext task to predict the missing patches from a masked input image. For an input image $\mathbf x$ and a mask $\mathbf m$ randomly drawn from a distribution, an encoder $f_\theta$ and a decoder $h_\theta$ are trained such that their composition $\mathcal I_\theta = h_\theta \circ f_\theta$ can reconstruct the original input $\mathbf x$ from the masked one, $\mathbf x^{\mathbf m} = \mathbf x \odot \mathbf m$ with the element-wise product $\odot$.

Then one adopts a Masked Auto-Encoder (MAE) \cite{he2022masked} reconstruction error
$$
\mathcal L_{mae}(\mathbf x^{\mathbf m}, \mathbf x ;\theta) = \mathcal D(\mathcal I_\theta(\mathbf x^{\mathbf m}), \mathbf x)
$$
with a distance measure $\mathcal D$ between reconstructed and original images to pre-train the network $\theta$.

\subsection{Direct Adversarial Approach}

Instance-wise perturbations can be easily extended to adversarially pretrain the network. Considering a constrained perturbation $\boldsymbol \delta$ to the masked image $\mathbf x^{\mathbf m}$, we have the following minimax problem for adversarial pre-training under the MIM framework
\begin{equation}\label{eq:mim_adv}
\min\limits_\theta\mathop\mathbb E\limits_{\mathbf x\sim\mathcal D}\max\limits_{\|\boldsymbol\delta\|_p\leq \epsilon}\mathcal L_{mae}(\mathbf x^{\mathbf m}+\boldsymbol\delta, \mathbf x;\theta).
\end{equation}
The perturbation is added onto the masked image that is fed into the network $\mathcal I_\theta$ to be adversarially pre-trained. Both FGSM \cite{goodfellow2014explaining} and PGD \cite{madry2017towards} approaches can be applied to solve the optimal perturbation $\boldsymbol\delta$ for the inner maximization problem.

\subsection{Adversarial Masking}
The above direct extension of instance-wise perturbations still relies on randomly drawn masks. An alternative method \cite{shi2022adversarial} considers to learn adversarial masks rather than the additive perturbations for the MIM-based pre-training.

It learns a mask generating function $\mathcal M_\phi$ that maps an input image $\mathbf x$ to a mask $\mathbf m$. The mask generating function is implemented with a network that gives real-valued entries on $[0,1]$ for an output mask. Then, the auto-encoder $\mathcal I_\theta$ and the mask generating network $\mathcal M_\phi$ are jointly trained by
\begin{equation}\label{eq:mim_mask}
\min\limits_\theta \max\limits_{\phi} \mathop\mathbb E\limits_{\mathbf x\sim\mathcal D}\mathcal D(\mathcal I_\theta(\mathbf x \odot \mathcal M_\phi(\mathbf x)), \mathbf x),
\end{equation}
where the loss is maximized over $\phi$ to adversarially mask the input image.

In \cite{shi2022adversarial}, multiple masks are generated for an input image $\mathbf x$. This is beneficial as different masks can train the auto-encoder network to reason about the relations between various parts of the image. However, a trivial solution may arise if a mask is all zeros (everything masked out) or all ones (nothing masked out). To avoid this problem, a sparsity penalty is imposed to penalize all-zero and all-one masks.

A variety of distance measures $\mathcal D$ have been considered in \cite{shi2022adversarial}. Defined on the image level, the distance can be the pixel-wise mean squared error as in MAE \cite{he2022masked}. Alternatively, the feature-level distance $\mathcal D(f_\theta(\mathbf x \odot \mathcal M_\phi(\mathbf x)), f_\theta(\mathbf x))$
 is also considered between the encoder outputs. In this case, to prevent the encoder outputs from collapsing to a constant, the contrastive loss and the BYOL loss can be adopted as the distance measure between original and recovered images \cite{shi2022adversarial}.

The adversarial masking in (\ref{eq:mim_mask}) can be viewed as a multiplicative adversary compared with the additive adversary in (\ref{eq:mim_adv}).

\subsection{Adversarial Positional Embeddings}

The Adversarial Positional Embeddings (AdPE) \cite{anonymous2022adpe} distorts the spatial structure of images by imposing adversarial perturbations on the positional embeddings in pretraining vision transformers. This seeks to prevent the transformer from learning trivial representations by merely using local correlations between tokens to predict missing patches. Instead, the transformer is forced to learn high-level representations in global contexts to infer the masked tokens as the local correlations are adversarially distorted. Both absolute \cite{shaw2018self} and relative positional embeddings \cite{wu2021rethinking} are considered on which adversarial perturbations are imposed. Moreover, the adversaries can be added in either embedding or coordinate mode. Specifically, in the embedding mode, the adversarial perturbations are added to the positional embeddings directly, while in the coordinate mode, they are added to the image coordinates to disturb the positional embeddings through a differentiable function. The latter is more direct in a way that distorts the coordinates to prevent the pretrained transformer from exploring the spatial correlations to predict missing patches.

The experiment results in \cite{anonymous2022adpe} demonstrate the adversarial perturbations on relative positional embeddings in coordinate mode perform the best among four variants. The formulation is as follows
$$
\min_{\boldsymbol\theta,\mathbf p}\max_{\|\boldsymbol\delta\|_q\leq\epsilon}\mathcal L_{MIM}(\{\mathbf p_{[\widetilde g(x_i-x_j+\delta_x),\widetilde g(y_i-y_j+\delta_y)]}|i\in\mathbb M\};\boldsymbol\theta)
$$
where $\mathcal L_{MIM}$ is the MIM loss such as the prediction error used in the MAE \cite{he2022masked} when reconstructing missing matches, $\mathbf p_{[u,v]}$ is the relative positional embedding with a 2D index of $[u,v]$, $\widetilde g(x_i-x_j+\delta_x)$ and $\widetilde g(y_i-y_j+\delta_y)$ are the continuous indexing function for $x-$ and $y-$coordinate respectively, and $\mathbb M$ is the index of masked patches. The spatial perturbation $\boldsymbol\delta$ is added on the relative coordinates along x- and y-axis, and the indexing function $\tilde g$ maps the perturbed relative coordinates onto real-valued indices to a dictionary of positional embeddings in $\{\mathbf p_{[u,v]}|u,v=-K,\cdots,K\}$ with bilinear interpolation \cite{anonymous2022adpe}. The optimal $\boldsymbol\delta$ is solved via the PGD with the network parameters $\boldsymbol \theta$ updated alternately by this minimax problem.



\section{Other Related Methods}\label{sec:other}

In addition to adversarial contrastive and MIM-based pretraining, there exist other works on  finding hard samples to boost the network pretraining, investigating adversaries with other pretraining methods, and exploring non-adversarial approaches for network robustness.  We review them in this section.

\subsection{Mining and Sampling Hard negatives}
Both instance-wise perturbations and memory-based adversarial negatives seek to find hard negatives to challenge the pre-training of deep networks so that more discriminative features can be learned to distinguish between positive and negative pairs. While both methods are derived by solving an optimization problem, other approaches such as directly mining \cite{kalantidis2020hard} or sampling \cite{robinson2020contrastive} hard negatives have also been studied.

{\noindent \bf Hard negative mining.} In \cite{kalantidis2020hard}, the top-$N$ similar negatives to a query are selected and mixed with the query and each other to generate harder negative samples. These harder negatives are appended to existing negative samples for the contrastive pre-training, yielding a hard negative mining approach.

{\noindent \bf Hard negative sampling.} In \cite{robinson2020contrastive}, a negative sampling distribution was formed to draw hard negatives that are close to a query anchor with different labels. Importance sampling was applied to sample such hard examples with a concentration hyperparameter $\beta$ controlling the hardness degree. The idea of debiased contrastive learning \cite{chuang2020debiased} from positive-unlabeled learning \cite{du2014analysis,elkan2008learning} was adopted to sample these hard negatives without access to true labels.

Hard negative mining and sampling approaches differ from the aforementioned adversarial contrastive pretraining as these hard negatives are not derived by maximizing the InfoNCE contrastive loss. Although these negative samples can also be viewed as adversarial in a sense that makes the pretraining task harder, they are not the worst-case adversarial examples as in typical adversarial approaches.


\subsection{Adversarial Methods Beyond Contrastive and MIM Pre-training}

The idea of adversarially pre-training deep networks goes beyond the CL and MIM frameworks. In particular, the instance-wise adversarial perturbations can easily be adapted to other self-supervised pre-training approaches such as BYOL \cite{grill2020bootstrap}, a self-supervised method by minimizing a mean-squared error between the representations of two transformed examples.

Formally in \cite{gowal2020self}, given a sample $\mathbf x$ along with its two augmented views $t(\mathbf x)$ and $t'(\mathbf x)$, they will go through an online and a target network, respectively, resulting in two embedded vectors $\mathbf z$ and $\mathbf z'$. The online and target networks share the same architecture, while the weights of the target network are an exponential moving average of its online counterpart.  A MLP predictor is added onto the online network to predict $\mathbf z'$ from $\mathbf z$. Putting the composition of the encoder, the projector and the predictor into $\kappa$, the mean-squared error of the BYOL objective can be written as
$$
\mathcal L_{byol}(t(\mathbf x), t'(\mathbf x);\theta)  = \left\|\dfrac{\kappa(t(\mathbf x))}{\|\kappa(t(\mathbf x))\|_2}-\dfrac{\mathbf z'}{\|\mathbf z'\|_2}\right\|_2^2
$$


An instance-wise attack is performed \cite{gowal2020self} by adding a constrained perturbation $\boldsymbol\delta$ to $t(\mathbf x)$,
$$
\arg\limits_{\boldsymbol\delta}\max\limits_{\|\boldsymbol\delta\|_p\leq \epsilon} \mathcal L_{byol}(t(\mathbf x)+\boldsymbol\delta, t'(\mathbf x);\theta).
$$
This problem can be solved iteratively as
$$
\boldsymbol\delta \leftarrow \Pi_{\|\boldsymbol\delta\|_p\leq \epsilon}[\boldsymbol \delta + \alpha \nabla_{\boldsymbol\delta} \mathcal L_{byol}(t(\mathbf x)+\boldsymbol\delta, t'(\mathbf x);\theta)]
$$
where $\alpha$ is the step size, and $\Pi$ is the projection onto the constraint perturbation.

With the optimal perturbation $\boldsymbol\delta^\star$, the BYOL loss is minimized to update the network weights as in other adversarial pretraining methods.

For other pretraining approaches, the instance-wise attacks can be formulated in the similar manner by maximizing the underlying loss to perturb on an instance input, followed alternately by minimizing the loss to update the network weights.

\subsection{Non-adversarial Methods for Adversarial Robustness}

Besides leveraging adversarial examples to make pre-trained networks robust to potential adversaries in downstream tasks, non-adversarial pretraining has also been developed without finding adversarial examples in the pretraining stage.

Inspired by randomized smoothing \cite{cohen2019certified,liu2018towards,cao2017mitigating,teng2019ell_1,zhang2019filling,levine2020robustness,yang2020randomized}, in~\cite{pang2022rush}, first one augmented view $t'(\mathbf x)$ of an input instance $\mathbf x$ is randomly smoothed via a distribution $\mathcal Q$. For example, one can adopt $\mathcal Q(\mathbf x)=\mathbf x+\boldsymbol \delta$, with an isotropic $d$-dimensional Gaussian distribution $\boldsymbol\delta\sim \mathcal N_d(0,\sigma^2)$ or a multi-dimensional uniform distribution $\boldsymbol\delta \sim U_d(-\mu,\mu)$.

Then, a simple non-adversarial pre-training objective attempts to pre-train a network parameterized with $\theta$ by minimizing a self-supervised loss $\mathcal L_{pre}$,
$$
\min\limits_{\theta} \mathop\mathbb E\limits_{t,t'\sim\mathcal T,\mathbf x\sim \mathcal D} \mathcal L_{pre}(t(\mathbf x),t'(\mathbf x);\theta)
$$
where $\mathcal L_{pre}$ can be chosen to any unsupervised objective such as contrastive loss based on SimCLR, and the BYOL loss minimizing the prediction error between online and target branches.

Once the network is pretrained, it is fine-tuned on downstream tasks, and the robustness of the fine-tuned network can be evaluated based on the adversarial examples resulting from a majority or an average voting \cite{pang2022rush}.

\section{Evaluation Protocols}\label{sec:eval}

After a deep network is adversarially pretrained, it needs to be adapted into a downstream task such as image classification to evaluate its performance on standard accuracy with clean data and/or robustness to various adversarial attacks.  While adversarial pretraining does not involve any labels, the pretrained network usually should be retrained with labeled examples for the downstream task. In literature, for the evaluation purpose the downstream retraining has been  performed in
various ways, which are discussed below.

{\noindent\bf Linear evaluation and robust linear evaluation.}  In self-supervised learning, the projector of a pretrained network is removed and replaced with a linear classifier. While the pretrained feature encoder is fixed, the linear classifier is trained with labeled examples. The linear classifier can be regularly trained by minimizing the cross-entropy loss with clean examples, or adversarially trained resulting in the robust linear evaluation as in \cite{kim2020adversarial}.

{\noindent\bf Supervised fine-tuning and adversarial fine-tuning.} Instead of training a linear classifier by fixing the pretrained encoder, the whole network can be supervisedly fine-tuned end-to-end merely by initializing the encoder with the adversarially pretrained model \cite{fan2021does}. Similarly, the supervised fine-tuning can be performed regularly with clean data or adversarially with adversarial examples.

{\noindent\bf Standard accuracy vs. robustness.} Standard accuracy is evaluated by applying clean examples and their groundtruth labels on the retrained model. This tests how well the model performs on regular data. On the contrary, the robustness is concerned with the adversarial attacks -- the retrained model is evaluated with adversarial examples generated with various attacking models, e.g., PGD attacks \cite{madry2017towards} with different types of norms ($\ell_\infty$, $\ell_1$, or $\ell_2$), their magnitudes, and the number of PGD steps. Black-box attacks generated from AT \cite{madry2017towards} and TRADES \cite{zhang2019theoretically} may also be used for the evaluation \cite{kim2020adversarial}.

\begin{table*}
\caption{Comparison between adversarial pretraining methods based on different evaluation protocols on CIRAR-10 and CIFAR-100. The results are reported in terms of both standard and robust accuracies based on $\ell_\infty$-attacks. AT denotes supervised adversarial training \cite{madry2017towards}, and SS denotes self-supervised loss used for fine-tuning \cite{kim2020adversarial}. * uses the linear evaluation based on voting strategy with randomized smoothing \cite{pang2022rush}. $\dag$ denotes the model pretrained on Image100-S and evaluated on downstream CIFAR-10 and CIFAR-100 datasets in the transfer learning setting with the base SimCLR included in the parenthesis for comparison \cite{shi2022adversarial}.}
\label{tab:adv}
\begin{center}
\setlength{\tabcolsep}{3.5mm}{
\subcaption{CIFAR-10}\begin{tabular}{lccc}
\toprule[1pt]
Method & Evaluation Protocol & Standard Accuracy&Robust Accuracy \\
\midrule[1pt]
AT \cite{madry2017towards} & adversarial fine-tuning & 78.99 & 47.41\\
TRADES \cite{zhang2019theoretically} & adversarial fine-tuning & 81.00 & 53.27\\
SimCLR \cite{kim2020adversarial} & linear evaluation & 91.25 & 0.63 \\
RoCL \cite{kim2020adversarial} & linear evaluation & 83.71 & 40.27\\
RoCL \cite{kim2020adversarial} & robust linear & 80.43 & 47.69\\
RoCL+AT+SS \cite{kim2020adversarial} & adversarial fine-tuning & 91.34&49.66\\
ACL \cite{jiang2020robust}& adversarial fine-tuning & 82.19 & 52.82\\
SimCLR+CLAE \cite{ho2020contrastive}& robust linear & 76.70$\pm$0.36 & - \\
ADVCL \cite{fan2021does} & linear evaluation & 81.35 & 51.00 \\
ADVCL \cite{fan2021does} & robust linear & 79.24 & 52.38\\
ADVCL \cite{fan2021does} & adversarial fine-tuning & 83.67 & 53.35\\
RUSH \cite{pang2022rush}& linear evaluation$^*$ &87.9&79.5\\
SimCLR+IFM \cite{robinson2021can}& linear evaluation & 91.9 & - \\
SimCLR+ADIOS \cite{shi2022adversarial} & linear evaluation & 34.6 (30.1)$^\dag$& -\\
SimCLR+ADIOS \cite{shi2022adversarial} & supervised fine-tuning & 93.4 (91.3)$^\dag$ & -\\
\bottomrule[1pt]
\end{tabular}}\\\vspace{2mm}
\setlength{\tabcolsep}{3.5mm}{
\subcaption{CIFAR-100}\begin{tabular}{lccc}
\toprule[1pt]
Method & Evaluation Protocol & Standard Accuracy&Robust Accuracy \\
\midrule[1pt]
AT \cite{madry2017towards}& adversarial fine-tuning & 49.49 & 23.00\\
TRADES \cite{zhang2019theoretically}& adversarial fine-tuning & 54.59 & 28.43\\
ACL \cite{jiang2020robust}& adversarial fine-tuning & 56.77 & 28.33\\
SimCLR+CLAE \cite{ho2020contrastive}& robust linear & 55.52$\pm$0.3 & - \\
ADVCL \cite{fan2021does} & linear evaluation & 47.98 & 27.99 \\
ADVCL \cite{fan2021does} & robust linear & 47.45 & 28.29\\
ADVCL \cite{fan2021does} & adversarial fine-tuning & 57.87 & 29.48 \\
RUSH \cite{pang2022rush}& linear evaluation$^*$ &57.1&46.6\\
SimCLR+IFM \cite{robinson2021can}& linear evaluation&68.8 & - \\
SimCLR+ADIOS \cite{shi2022adversarial} & linear evaluation & 11.0 (10.2)$^\dag$& -\\
SimCLR+ADIOS \cite{shi2022adversarial} & supervised fine-tuning & 71.8 (70.0)$^\dag$ & -\\
\bottomrule[1pt]
\end{tabular}
}

\end{center}
\end{table*}

\begin{table}[t]
\caption{Comparison of CL- and MIM-pretrained model on ViT-B and ViT-L. The supervised fine-tuning is performed with Imagenet labels in terms of the top-1 on the Imagenet1k dataset. Among them are the adversarial approaches including FGSM and AdPE. By comparison, the instance-wise adversarial model -- FGSM does not improve the top-1 accuracy compared to its base model MAE+, while the AdPE performs better than MAE+.}
\label{tab:advall}
\begin{center}

\setlength{\tabcolsep}{1mm}{
\begin{tabular}{lccccccc}
\toprule[1.5pt]
Method & Type&Adversarial &Extra Model &Epochs&ViT-B& ViT-L \\
\midrule[1.5pt]
supervised~\citep{he2022masked}& Supervised &no&-&300&82.3&82.6\\
MoCo-v3~\citep{chen2021empirical}&Contrastive &no&momentum ViT & 300&83.2&84.1\\
DINO~\citep{caron2021emerging}&Contrastive &no&momentum ViT&300&82.8&-\\
iBOT~\citep{zhou2021ibot}&Contrastive+MIM &no& momentum ViT&1600& 84.0&84.8\\
BEiT~\citep{bao2021beit}&MIM &no&DALLE+dVAE & 800& 83.2&85.2\\
data2vec~\citep{baevski2022data2vec}&Contrastive&no& momentum ViT&800&84.2&86.2\\
CAE~\citep{chen2022context} & MIM &no&DALLE tokenizer&1600&83.9&86.3\\

\midrule
SimMIM~\citep{xie2022simmim} &MIM&no&-&800&83.8&85.4\\
MaskFeat~\citep{wei2022masked} &MIM&no&-&1600&84.0&85.7 \\
MAE~\citep{he2022masked}&MIM &no&-&1600& 83.6&85.9 \\
MAE+~\cite{anonymous2022adpe} & MIM &no& - & 400 & 83.51 & - \\
MAE+~\cite{anonymous2022adpe} & MIM &no& - &1600& 83.9 & 86.0\\
FGSM~\cite{goodfellow2014explaining} & MIM &yes& - & 400 & 83.09 & - \\
AdPE~\cite{anonymous2022adpe} &MIM  &yes& - &1600 & { 84.4} &{ 86.3}\\
\bottomrule[1.5pt]
\end{tabular}}
\end{center}
\end{table}

The large variety of objectives and evaluation protocols make it complicated to compare different adversarial pretraining models.  For example, most of instance-wise adversarial pretraining models \cite{kim2020adversarial,ho2020contrastive,jiang2020robust,robinson2021can,gowal2020self,pang2022rush} are intentionally developed to be robust against instance-wise attacks, and are thus evaluated based on the robust accuracy. Meanwhile, they are also assessed in standard accuracy on downstream tasks with clean examples.  Table~\ref{tab:adv} summarizes some reported results in terms of different evaluations protocols. Most of evaluations are performed on CIFAR-10 \cite{krizhevsky2009learning}, CIFAR-100 \cite{krizhevsky2009learning}, and STL-10 \cite{coates2011analysis}. Standard and robust accuracies are often at odds with one another. This is not surprising as these models are developed purposely in defence of such attacks, and a trade-off usually exists to sacrifice standard accuracy for robustness against various attacks.  This is also aligned with the findings from adversarial training \cite{tsipras2018robustness}.

On the contrary, the {\em feature-level adversarial pretraining} approaches such as memory-based adversarial contrastive learning \cite{hu2021adco,wang2022caco}, hard negative mining and sampling \cite{
robinson2020contrastive,kalantidis2020hard} and MIM-based adversarial pretraining of transformers \cite{anonymous2022adpe,shi2022adversarial} are often more concerned about the generalization ability measured by their accuracies on downstream tasks, since these methods are adversarially pre-trained to avoid learning trivial features without being instance-wise attacked. The pretrained network is fine-tuned for downstream tasks without adversarial examples, and the fine-tuned network is evaluated on clean test examples. As these models are concerned with the improvements over the other state-of-the-art self-supervised methods, for the sake of a fair comparison, they are usually evaluated on Imagenet1K \cite{imagenet_cvpr09}. Table~\ref{tab:advall} and Table~\ref{tab:imagenet800} compares the top-1 accuracy of various self-supervised pretraining methods on Imagenet1K with vision transformers (ViT-B and ViT-L) and ResNet-50 backbones. The results show that the adversarially pretrained networks can improve the accuracy over fewer pretraining epochs, demonstrating their efficiency in exploring adversarial examples to learn more discriminative features generalizable to future tasks.

\section{Emerging Trends and Future Directions}\label{sec:emerging}
In this section, we will reveal and discuss several aspects of emerging trends and potential future directions.

\subsection{A Hybrid Model: Adversarial vs. Cooperative Pretraining}

\begin{figure*}[t]
    \centering
    \begin{subfigure}[c]{1.0\textwidth}
        \includegraphics[width=\textwidth]{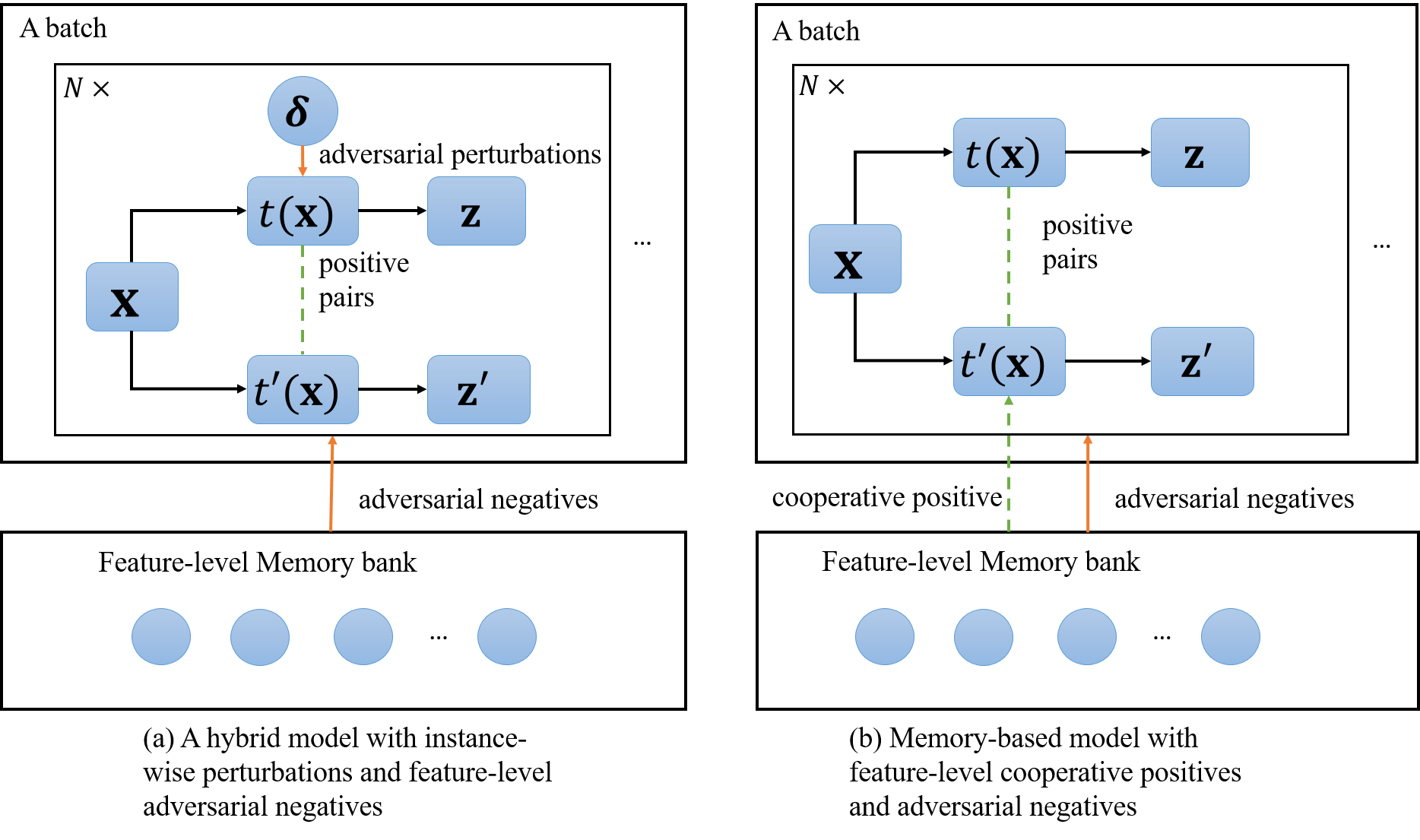}
    \end{subfigure}\\
    ~ 
    \caption{ (a) A hybrid model with model-free instance-wise perturbations and model-based adversarial negatives. Instead of sticking with a positive sample from the memory bank, a hybrid model will apply instance-wise perturbation to each query $t(\mathbf x)$. This combines the memory-free instance-wise perturbations and memory-based adversarial negatives. Both  perturbations and negatives are adversarially learned. (b) Memory-based model with cooperative positives and adversarial negatives. For each query sample $t(\mathbf x)$, a sample is picked from the memory bank as its positive counterpart. This positive sample chosen from the memory bank is learned cooperatively rather than adversarially by minimizing the contrastive loss.}\label{fig:cooperative}
\end{figure*}

\begin{table}
\caption{Comparison between adversarial pretraining with cooperative positives and other methods \cite{wang2022caco}. Top-1 accuracy under the linear evaluation on Imagenet1K with the ResNet-50 backbone. The table compares the methods using a single crop augmentation pre-trained with various epochs. As mentioned in the introduction of Section~\ref{sec:intro}, these pretraining approaches are learning on feature level adversaries (if any) rather than upon instance-wise perturbations, and thus they aim to learn more generalizable representations instead of those robust against certain type of instance-level attacks.}
\label{tab:imagenet800}
\begin{center}
\setlength{\tabcolsep}{3.5mm}{
\begin{tabular}{lccccc}
\toprule[1pt]
Method & epochs & batch size&top-1&adversarial&cooperative \\
\midrule[1pt]
InstDisc~\cite{wu2018unsupervised} & 200 & 256  & 58.5 & no & no \\
MoCHi \cite{kalantidis2020hard} &200& 512 & 68.0 & yes & no\\
MoCHi \cite{kalantidis2020hard} & 800 & 512 & 68.7 & yes & no\\
SeLa~\cite{asano2019self} &400& 256  & 61.5 & no & no\\
PIRL~\cite{misra2020self}&800& 1024 & 63.6 &no & no\\
CMC~\cite{tian2019contrastive} &240& 128  & 66.2& no & no\\
SimCLR~\cite{chen2020simple} & 200 & 256  & 61.9 & no & no\\
SimCLR~\cite{chen2020simple} &800& 4096 & 69.3 & no & no\\
PIC~\cite{cao2020parametric} & 200&512 & 67.6 & no & no\\
PIC~\cite{cao2020parametric} &1600& 512 &  70.8 & no & no\\
CPC v2~\cite{henaff2019data} & 200&512  & 63.8 & no & no\\
PCL v2~\cite{li2020prototypical} & 200&256  & 67.6 & no & no\\
MoCo v2~\cite{chen2020improved} & 200&256  & 67.5 & no & no\\
MoCo v2~\cite{chen2020improved} &800& 256 &  71.1 & no & no\\
MoCo v3~\cite{chen2021empirical}&1000&4096&74.6 & no & no\\
InfoMin Aug~\cite{tian2020makes}&800& 256 &  73.0 & no & no\\
SimSiam~\cite{chen2021exploring} &200&256&70.0 & no & no\\
SimSiam~\cite{chen2021exploring} &800&256&71.3 & no & no\\
SWAV~\cite{caron2020unsupervised} & 200&4096  & 69.1 & no & no\\
SWAV~\cite{caron2020unsupervised} &800& 4096  & 71.8 & no & no\\
NNCLR~\cite{dwibedi2021little} &200&4096&70.7 & no & no\\
NNCLR~\cite{dwibedi2021little} &800&4096&74.9 & no & no\\
BYOL~\cite{grill2020bootstrap} & 200&4096 &70.6 & no & no\\
BYOL~\cite{grill2020bootstrap}&1000& 4096 & 74.3 & no & no\\
AdCo~\cite{hu2021adco} &200&256  & 68.2 & yes & no\\
AdCo~\cite{hu2021adco}&800 &256  & 72.8 & yes & no\\
CaCo~\cite{wang2022caco} &200&1024 &{ 71.3} & yes & yes\\
CaCo~\cite{wang2022caco} &800&4096& { 75.4} & yes & yes\\
\bottomrule[1pt]
\end{tabular}}
\end{center}
\end{table}

In Section~\ref{sec:mem}, we discussed how adversarial contrastive pretraining is applied in \cite{hu2021adco} to generate adversarial negatives. However, an unanswered question there is whether and how to perform adversaries against positives. As shown in Eq.~(\ref{eq:mfree_neg_update}), the adversaries are only performed on the shared negatives. This is in contrast to instance-wise attacks in Section~\ref{sec:mfree}, where the positives are also perturbed adversarially to pre-train the deep network.

There is a natural extension by combinng such shared negative adversaries in AdCo \cite{hu2021adco} and the instance-wise attacks on positives. Given a pair of positives $t(\mathbf x)$ and $t'(\mathbf x)$ transformed from the same instance $\mathbf x$ with $t,t'\sim\mathcal T$, we can still apply an instance-wise attack $\boldsymbol \delta$ against $t(\mathbf x)$, leading to a instance-wise adversarial positive as in Eq.~(\ref{eq:mfree_delta}). Combining the perturbed positive by $\boldsymbol \delta$  on  $t(\mathbf x)$ with the AdCo,  it yields a hybrid model with instance-wise perturbations and memory-based adversarial negatives.

Alternatively, if we wish to stick with the memory-based model for positive samples,  it turns out that we need to abandon  adversarial positives and instead switch to cooperative ones \cite{wang2022caco}.  This is not hard to understand -- with all positives in the memory bank shared among query anchors in a batch, adversarial updates to these positives will push them away from these query anchors . This will  push these positives into a region of feature space that is far away from the representations of query anchors as for shared adversarial negatives in AdCo.  It will eventually make the network pre-training meaningless if the positives become completely irrelevant to the learned representations of incoming query samples.

Figure~\ref{fig:cooperative} compares the memory-based model with cooperative positives and the aforementioned hybrid model. Indeed, \cite{wang2022caco} studied this problem and revealed that the positives shared in a memory bank ought to be learned cooperatively rather than adversarially, that is
\begin{equation}\label{eq:co_pos}
\mathbf z_{pos} \leftarrow \mathbf z_{pos} + \dfrac{1}{\tau}[1-p(\mathbf z_{pos}|\mathbf z)] \mathbf z,
\end{equation}
where $\mathbf z_{pos}$ is a  positive chosen from the memory bank for a query anchor $\mathbf z$. Table~\ref{tab:imagenet800} compares such a cooperative positive model with the other adversarial pre-training results based on the linear evaluation protocol under various batch sizes and pre-training epochs.

On the contrary, adversarial updates to memory-bank positives would lead to trivial results in memory-based model. For example,
if the cooperative positives in (\ref{eq:co_pos}) were changed to the following adversarial formulation
\begin{equation}
\mathbf z_{pos} \leftarrow \mathbf z_{pos} - \dfrac{1}{\tau}[1-p(\mathbf z_{pos}|\mathbf z)] \mathbf z,
\end{equation}
the network pre-training would collapse to trivial representations, as evidenced by a very low accuracy in downstream tasks close to zero \cite{wang2022caco}.

Now we have an open problem to study in future about the essential connection between instance-wise perturbations and memory-based adversarial model  --  {\em can we still perform adversarial attacks against shared positives without over-memorizing them in the memory bank while avoiding the trivial results?  How much advantage can we obtain from cooperative positives over adversarial perturbations, or vice versa?} These are open questions for future research.



\subsection{Unifying Adversarial Contrastive Learning and Masked Auto-Encoders}
In this paper, we have discussed the adversarial pretraining based on contrastive learning and MIM, respectively.
Existing works \cite{zhou2021ibot,huang2022contrastive} have demonstrated that by combining the objectives for contrastive learning and MIM for pretraining vision transformers, more generalizable representations can be learned.

Essentially, the contrastive loss is designed as a classification objective of distinguishing between the augmented views of different instances. On the other hand, the MIM objective is related with a generative task of reconstructing masked image patches by exploring the relations with unmasked contexts. Combing two objectives is expected to learn both discriminative and generative features that well generalize to future tasks.

There are evidences showing the MIM representations perform well when it is fine-tuned end-to-end to a downstream task, while the contrastively pretrained features are better in linear evaluations by training a linear classifier by freezing the pretrained network.  Combing the contrastive pretraining with the MIM objective further could improve the generalization accuracy in downstream tasks \cite{zhou2021ibot,huang2022contrastive}.

However, no effort has been made to explore the adversarial pretraining along this direction. A straight idea is to find the worst-case perturbations by jointly maximizing the contrastive loss and the MAE loss of a MIM task, and to use the obtained adversarial perturbations to train a robust representation by minimizing the joint loss.  {\em Then, will this lead to improved generalization accuracy and/or robustness to adversarial attacks?  Are there any other ways to pre-train both discriminative and generative representations that robustly generalize to the future applications?} These questions deserve further studies in future.

\subsection{Accuracy vs. Robustness}

The dual goal of adversarial pre-training is to achieve better results in both standard accuracy and robustness against adversarial attacks. The standard accuracy is related to the generalization ability of pre-trained representations to downstream applications that usually have a different objective from the pretext task for pretraining. The robustness is usually concerned with adversarial attacks derived from the downstream tasks. For example, one uses contrastive pretraining to learn an unsupervised representation with unlabeled data, and it is then fine-tuned to a classification task. In this case, we are concerned with the classification accuracy on the test set, as well as its robustness against adversarial examples.

Empirical evidences have shown that a high robustness against worst-case adversarial examples often comes with a low standard accuracy \cite{tsipras2018robustness}. In particular, such an accuracy-robustness dilemma is prevailing in instance-wise perturbations \cite{kim2020adversarial,ho2020contrastive,jiang2020robust,robinson2021can,gowal2020self,pang2022rush}, since these models are designed to be pre-trained against adversarially perturbed examples --  they are trained to be invariant to these perturbation noises.

On the contrary, memory-based adversarial pre-training \cite{hu2021adco,wang2022caco} abandons the direct perturbations on individual instances, and instead learns shared adversarial negatives that are hard to be distinguished from their positive counterparts. It thus focuses on learning more discriminative features that can yield better generalization performance on downstream tasks. For example, Table~\ref{tab:imagenet800} shows these memory-based adversarial pre-training has competitive results over the other instance-wise pre-training models.

Therefore, opens questions are: {\em is it an intrinsic nature that one must trade generalizability for robustness, or vice versa? Is it possible for us to combine the advantages of memory-free instance-wise attacks and memory-base adversarial pretraining to achieve both high accuracy and strong robustness?} Answering these questions demands tremendous efforts from both theoretical and practical perspectives.

\section{Conclusion}\label{sec:concl}
This paper reviews the adversarial pretraining by grouping related works into memory-free instance-wise attacks and memory-based adversaries employing both raw input and features. While the former imposes sample-specific perturbations on individual examples, the latter applies a shared memory of adversaries during the model pretraining.  We review the adversarial pretraining methods aligned with Contrastive Learning (CL), Masked Image Modeling (MIM) and beyond.  The issues related to computing overheads, input and feature level adversaries, and a hybrid objective combining adversarial and standard losses are also reviewed.  Emerging trends lead to several open problems about adversarial or cooperative positives in contrastive pretraining, combining contrastive and MIM-based adversaries, and the intrinsic trade-off between accuracy and robustness for adversarial pretraining. The answers to these open problems need tremendous efforts in order to to reveal the connections between the pretext task for adversarial pretraining and the trade-off or unification of the generalizability and robustness in downstream tasks.


\bibliographystyle{ACM-Reference-Format}
\bibliography{transformers,pretraining,AdCo}

%
%
%
%
%
%
%
%

\end{document}